\documentclass[conference]{IEEEtran}

\IEEEoverridecommandlockouts                              

\usepackage{comment}
\usepackage{graphicx}
\usepackage{caption}
\usepackage{subcaption}

\usepackage{cite}
\usepackage{wrapfig}

\usepackage{algorithm}
\usepackage{algorithmic}
\usepackage{bibentry}
\usepackage[utf8]{inputenc}
\usepackage{float}
\usepackage{multicol}
\usepackage{multirow}

\usepackage{amsmath}
\newcommand{\algorithmicinput}{\textbf{\textsl{Input.}}}

\newcommand{\INPUT}{\item[\algorithmicinput]}

\usepackage{multicol}
\usepackage{breqn}
\usepackage{amsfonts} 

\title{Predicting Sample Collision with Neural Networks}
\author{Tuan Tran$^{1}$, Jory Denny$^{2}$, and Chinwe Ekenna$^{1}$
\thanks{*This work was not supported by any organization.}
\thanks{$^{1}$Tuan Tran and Chinwe Ekenna are with the Department of Computer Science, University at Albany, SUNY, NY 12206, USA {\tt\small\{ttran3, cekenna\}@albany.edu}.}%
\thanks{$^{2}$Jory Denny is with the Department of Mathematics and Computer Science, University of Richmond, VA 23173, USA {\tt\small jdenny@richmond.edu}.}%
}

\date{July 2019}

\begin{document}

\maketitle


\begin{abstract}
Many state-of-art robotics applications require fast and efficient motion planning algorithms. 
Existing motion planning methods become less effective as the dimensionality of the robot and its workspace increases, especially the computational cost of collision detection routines. 
In this work, we present a framework to address the cost of expensive primitive operations in sampling-based motion planning. 
This framework determines the validity of a sample robot configuration through a novel combination of a Contractive AutoEncoder (CAE), which captures a occupancy grids representation of the robot's workspace, and a Multilayer Perceptron, which efficiently predicts the collision state of the robot from the CAE and the robot's configuration. 
We evaluate our framework on multiple planning problems with a variety of robots in 2D and 3D workspaces. 
The results show that (1) the framework is computationally efficient in all investigated problems, and (2) the framework generalizes well to new workspaces.
\end{abstract}



\section{Introduction}
	
Recently, machine learning has been infused in many solutions to notoriously difficult problems in robotics. These strategies have seen a wide array of success in visual sensing\cite{espiau1992new}, task learning\cite{nehaniv2007imitation}, and human-robot cooperation\cite{trautman2015robot}, etc. It is important to understand include learning in all levels of a robot's computation stack including common sub-components to solving these problems, e.g., subroutines for solving the motion planning problem.

In this work, we apply machine learning to predict the validity of robot configurations, a common procedure in sampling-based motion planners. Sampling-based approaches have seen broad applicability to many high-dimensional and complex motion planning problems in robotics\cite{denny2013lazy}, computer graphics\cite{kobbelt2004survey}, and computational biology\cite{ekenna2016adaptive}. Most of these approaches rely on the classification of robotic configurations into valid, e.g., collision-free, and invalid samples, while they work to construct graph approximations of a state space, called a roadmap. Thus, these approaches avoid full representation of the topology of the state space. However, as the dimensionality of the robot and the complexity of the workspace increase, the primitive operations of sampling-based methods become computationally obstructive\cite{bialkowski2013efficient}.

Prior research has focused on improved geometric analysis techniques for collision checking\cite{lin1998collision}, approximately modeling state space obstacles to predict the collision status of robot configurations, and lazily invoking the configuration validation subroutine\cite{denny2013lazy}, among others. While these methods have shown success in many applications, none have fully and permanently bounded the computational cost of validating configurations in sampling-based motion planners.

\begin{figure*}[htb!]
 \centering
 \begin{subfigure}{0.25\textwidth}
 \centering
 \includegraphics[width=\textwidth,height=100pt]{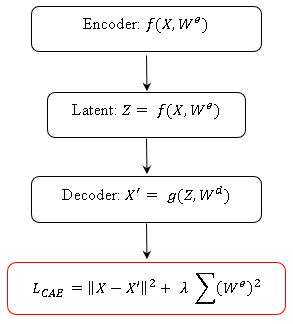}
   \caption{Offline: Contractive AutoEncoder}
 \end{subfigure}
 \hspace{1cm}
 \begin{subfigure}{0.25\textwidth}
 \centering
 \includegraphics[width=\textwidth,height=100pt]{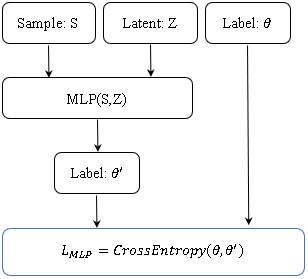}
  \caption{Offline: Multilayer Perceptron}
 \end{subfigure}
 \hspace{1cm}
 \begin{subfigure}{0.25\textwidth}
 \centering
 \includegraphics[width=\textwidth,height=100pt]{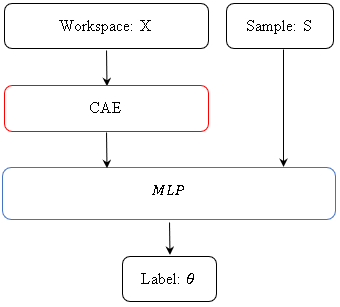}
  \caption{Online: Sample Validation}
 \end{subfigure}
 \caption{Overview of Proposed Framework}
 \label{fig:framework}
 \vspace{-4mm}
\end{figure*}

We propose a novel framework to efficiently and effectively predict the validity of robot configurations in complex motion planning problems. Figure \ref{fig:framework} shows the two parts of our framework: (1) Offline training phase: a Contractive AutoEncoder (CAE)\cite{rifai2011contractive}, is trained to embed the robot's workspace into a latent space, and a Multilayer Perceptron (MLP), is trained to predict the validity of a robots' configurations given the CAE encoding of the workspace; (2) Online sampling validation phase: a workspace is encoded into latent space by CAE and its samples are validated by MLP using that encoded workspace. We examine the potential impact of such an approach with a variety of robots in complex 2D and 3D workspaces.
For this work, we mainly focus on sampling because finding valid samples becomes paramount in narrow and cluttered passages. This work presents a first step towards applying our unique autoencoder variant to Sampling-based Motion Planning (SBMP).
Our results indicate that: our framework is computationally efficient in all investigated problems, and our approach generalizes well to previously un-encoded workspaces.
As such, our framework can be applied to many existing sampling-based routines to improve their computational efficiency.

\section{RELATED WORK}

\subsection{Motion Planning Primitives}

Let $X \subset \mathbb{R}^d$ denote a given state space, where $d \in \mathbb{N}$ is the dimensionality of the state space. 
The obstacle and obstacle-free state spaces are defined as $X_{obs} \subset X$ and $X_{free} = X \backslash X_{obs}$ respectively. 
A path $\tau$ for the robot is defined as a series of states $\{x_0, ..., x_n\}$, where n is the number of time steps in the path. A path is said to be feasible if it lies entirely in the obstacle-free space $X_{free}$

The motion planning problem is defined as: given a state space $X = X_{free}\subset X_{obs}$, an initial state $x_{init} \in X_{free}$, and a goal region  $X_{goal} \subseteq X_{free}$, find a feasible path $\tau \in X_{free}$ such that $\tau_{start} = x_{init}$ and $\tau_{end} \in X_{goal}$. If no such path exists, report failure.

\subsection{Sampling-based Motion Planning (SBMP)}

A common solution to the motion planning problem is the utilization of sampling-based methods. Most often, sampling-based motion planners randomly select and progressively expand and connect robot configurations to form approximate graph representations of $X_{free}$, called roadmaps. A roadmap encodes valid states as its nodes and transitions between them as its edges. In order to find a path, a starting state and goal region are connected to the roadmap, and a feasible path is extracted from it.

These techniques employ the invocation of a "black box" collision detection module that classifies any state in either $X_{free}$ or $X_{obs}$ to avoid explicit construction of $X_{obs}$. This is used in the verification of both the nodes and the edges of any roadmap constructed using these methodologies. Commonly sampling-based strategies offer guarantees on probabilistic completeness and asymptotic optimality \cite{ichter2018robot}.

The main paradigms of sampling-based motion planning are the Probabilistic RoadMap \cite{kavraki1994probabilistic} and the Rapidly-exploring Random Tree (RRT) algorithms \cite{lavalle1998rapidly}. Both PRM and RRT are known to be probabilistically complete, i.e., if a (robust) solution exists, then a solution will almost surely be found as the number of samples increases \cite{bialkowski2013efficient}.
Many proposed variations on PRM and RRT improve their performance for a variety of scenarios. Inoue et al.\cite{inoue2019robot} proposed a robot path planning method that combines RRT and long short-term memory network, which recalls the path of a robot by training with a large number of paths generated by RRT. 
Zhang et.al.\cite{zhang2018path} improved RRT algorithm by using a target bias sampling strategy, considering both distance and rotation angle when choosing the nearest neighbor, and making the planned path as smooth as possible using radial basis function neural network. 
Hauser proposed Lazy-PRM$^*$\cite{hauser2015lazy} a lazy collision checking strategy, which avoids checking connections that have no chance to improve upon the current best path, thus reducing the per-sample computational cost and accelerating solution convergence. 
This work likewise aims to improve the performance of sampling-based approaches --- specifically through reducing the cost of collision detection routines.

\subsection {Neural Network in Motion Planning}

Representation learning~\cite{bengio2013representation} mainly aims to extract features from unstructured data, to either achieve a lower dimensional representation (often referred to as encoding) or learn features for supervised learning or reinforcement learning. 
There have been some recent works to utilize a learned low-dimensional latent model for motion planning. Ichter et al.~\cite{ichter2018learning} and Zhang et al~\cite{zhang2018learning} used the Conditional Variational Autoencoders to learn a non-uniform sampling methodology of a sampling-based motion planning algorithm. The work in \cite{ichter2019robot} provides a new SBMP method using the learned latent space and plan motions directly in that space. Moreover, Chen et al.~\cite{chen2016dynamic} proposed a time-dependent variational autoencoder to learn dynamic motion primitives for humanoids, which could be applied to sampling configurations in motion planning.
Qureshi et al.~\cite{qureshi2018motion} 
learned a low-dimensional representation of the obstacle space by using a Contractive Autoencoder to encode a point cloud of the obstacles into a latent space and then used a feed-forward neural network to predict the next step an optimal planner would take given a start and goal.
In this work, we aim to use a lower dimensional representation to reduce the cost of collision detection for use in motion planning algorithms.

The neural network is becoming prominent in path planning for its outstanding non-linear mapping ability. The shortcomings of neural network models include hardware requirements to reduce training time and difficulty of obtaining the best parameters for learning. 
However, with the fast development of hardware computing capabilities and undemanding requirements of path planning, those shortcomings are fading steadily. 
Li et al.~\cite{li2018neural} addressed a motion planning problem for vehicles with non-linear dynamics in a clustered environment using near-optimal RRT, which utilizes a neural network to estimate the cost function considering non-linear kinodynamic constraints. Moreover, Shi et al.~\cite{shi2017motion} presented a motion planning system based on hybrid deep learning, which uses a convolutional neural network to reduce the dimension of the input image, a recurrent neural network to create a path tracking model, and a fully connected neural network to construct a control model. Chen et al.~\cite{chen2015optimal} proposed an optimal robot path planning system that builds a map, plans optimal paths, and maneuvers mobile robots, in which a simplified neural network is used to calculate the optimal trajectory for the robot.
In this work, we make use of the multilayer perceptron to quickly and efficiently determine the validity of generated samples for the motion planning problem.




\section{RESEARCH FRAMEWORK}

This section introduces our proposed framework. It is a neural network based motion planner comprised of two phases. The first phase trains a Contractive AutoEncoder and a Multilayer Perceptrons in an offline, apriori fashion. The second phase performs path finding computations online using the trained CAE and MLP.

\subsection{Offline Training}

The proposed framework uses two neural network models to solve the motion planning problem. 
The first is a CAE that embeds a representation of the workspace, corresponding to a point cloud, into a latent space. 
The second is a MLP which predicts the validity of a given input sample and the CAE encoding of the workspace.
Thus, instead of using the information of the whole workspace to decide whether a sample is valid or invalid, only the limited information from the encoded latent space is used.

\subsubsection{Contractive AutoEncoder (CAE)}

The standard construction of a CAE consists of training neural networks for the encoder and the decoder. The encoder and the decoder are trained together using the reconstruction error. The output of the encoder represents a reduced representation of the initial input, and the decoder reconstructs that initial input from an encoded representation by minimizing a cost or loss function.

A CAE is used to embed workspace occupancy grids $X$ into an invariant and robust latent space 
$Z 	\subseteq \mathbb{R}^m$, where $m \in \mathbb{N}$ is the dimensionality of the latent space. Let $f(x, W^e)$ be an encoding function with weight matrix $W^e$ that encodes an input vector $x \in X$ into a vector in the latent space $z \in Z$. A decoding function $g(z, W^d)$, with weight matrix $W^d$, decodes a vector from the latent space $z \in Z$ back into a vector in the workspace $x \in X$. The
objective function (loss function) for the CAE is:

\begin{equation}
L_{CAE} = \frac{1}{|D|} \sum_{x \in D} ||x - g(f(x,W^e),W^d)||^2 + \lambda \sum_{ij} (W_{ij}^e)^2
\end{equation}
 
where $\lambda$ is a penalizing coefficient and $D \subseteq X $ is the occupancy grid data for different workspaces.
The penalizing term forces the latent space $z = f(x, W^e)$ to be contractive in the neighborhood of the training data which results in an invariant and robust feature learning~\cite{rifai2011contractive}.
Since we use linear encoder, the loss function of CAE is similar to the loss function of a regularized autoencoder. Moreover, our training data mainly consists of an unlabeled representation of the environment, we apply autoencoders to learn the low-dimensional representation of the training data instead of end-to-end architecture. By using the reconstruction (error) from the decoder, we could analyze a quantifiable metric used to decide which low-dimensional representation accurately replicates our environment the best.

\subsubsection{Multilayer Perceptrons (MLP)}
We use a multilayer perceptron to perform sample validity. Given a workspace encoding $z = f(x,W^e) \in Z$, the samples' information, MLP predicts whether a sample is valid or not.
The training objective for the MLP is to minimize the classification loss between the predicted sample's label and its actual label. Thus we use Cross-Entropy as our loss function since it's been shown to perform excellently for classification problems \cite{covington2016deep,sainath2012auto}.

\subsection{Online Sample Validation and Path Planning}
The online phase exploits the neural models from the offline
phase to do motion planning in cluttered and complex workspaces.
Algorithm \ref{algo} presents the overall path generation procedure. 
First, using the offline-trained CAE, $W$ is encoded into a latent space $Z$. Then, a set of samples $S$ is generated based on the specification of a robot $R$ and its particular workspace $W$. 
Next, the offline-trained MLP takes the sample set $S$ and the encoded latent space $Z$ as input to determine a set of probably valid samples $S' \subseteq S$.
Then, the probably valid sample set $S'$ is connected to create a roadmap $R$ by a proximity search function. If $R$ is not found, additional valid samples would be generated using auxiliary samplers. Finally, from $R$, a local planning function would return a feasible path between a given start and goal.
We will discuss in more detail for each step in our algorithm in the following sections.

\begin{algorithm}
\caption{SBMP Variant}\label{algo}
\begin{algorithmic}[1]
    \INPUT{New workspace $X$ (point cloud data), Query $x_{init}$, $X_{goal}$}
    \STATE Use the CAE to encode $X$ into the latent space $Z$.
    \STATE Randomly sample a set of states, $S$. 
   	\STATE Feed $S$ and $Z$ into MLP, which predicts the validity of each sample in $S$. After this step $S'$ a set of most probably valid samples are retained.
	\STATE Feed $S'$ into a standard PRM approach to yield a valid roadmap $R$. If $R$ is not found, create additional valid samples using auxiliary samplers. 
	\STATE Extract a path $\tau$ from $R$ between $x_{init}$ and $X_{goal}$.
\end{algorithmic}
\end{algorithm}

\subsubsection{Sample Generation}
We use the following auxiliary samplers to generate the initial sample set $S$ from each workspace. Basic PRM (BS)\cite{kavraki1994probabilistic} creates samples uniformly and randomly. Obstacle Based PRM (OB)\cite{amato1998obprm} creates samples near the boundary of obstacles. Gaussian Sampler (G)\cite{boor1999gaussian} also creates samples around obstacles using adaptive probability and collision data. Bridge Test (BT)\cite{hsu2003bridge} creates samples using a short segment with two configurations and their midpoint.

\subsubsection{Workspace Encoder}

The encoding function $f(x, W^e)$, trained in the offline phase, is used to map the workspace occupancy grid  $x \in X$ into a latent space $Z \subseteq \mathbb{R}^m$.

\subsubsection{Sample's Validity}
The MLP, a feed-forward neural network from the offline phase, takes the workspace encoding $Z$, the samples' information $S$, and predict samples' validity to create $S'$. In short, MLP is a classifier distinguishing between valid and invalid samples. To introduce stochasticity into the MLP and to prevent over-fitting, some of the hidden units in each hidden layer of the MLP were dropped out with a probability $p$. Dropout is applied layer-wise to a neural network and it drops each unit in the hidden layer with a probability $p : [0, 1] \in \mathbb{R}$. 

\subsubsection{Connection and Path Planning}
We use a straight line local planner. Particularly a PRM approach is applied to $S'$ to generate a roadmap $R$, and the path is extracted and verified.


\section{IMPLEMENTATION DETAILS}

This section gives the implementation details of our framework.
The proposed neural models, CAE and MLP, are implemented in PyTorch\cite{paszke2017automatic}. 

\subsection{Data Collection}

\subsubsection{Workspace Data}

\begin{figure}[!tbp]
  \centering
  \begin{minipage}[b]{0.3\textwidth}
    \centering
    \includegraphics[width=\textwidth, height = 70pt]{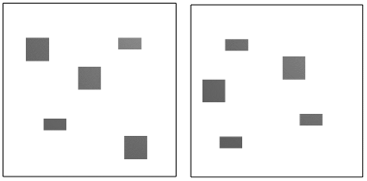} 
    \caption{Examples of 2D simple workspace}
    \label{fig:2D} 
    \vspace{-4mm}
  \end{minipage}
  \hfill
  \begin{minipage}[b]{0.3\textwidth}
    \centering
    \includegraphics[width=\textwidth, height = 70pt]{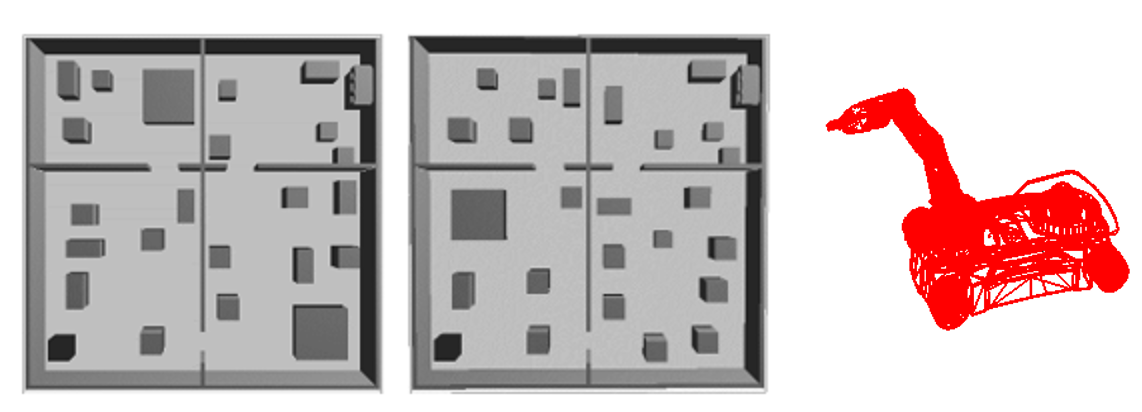} 
    \caption{Examples of 3D office-like workspace with Kuka robot}
    \label{fig:3DO} 
  \end{minipage}
  \hfill
  \begin{minipage}[b]{0.3\textwidth}
    \centering
    \includegraphics[width=90pt, height = 70pt]{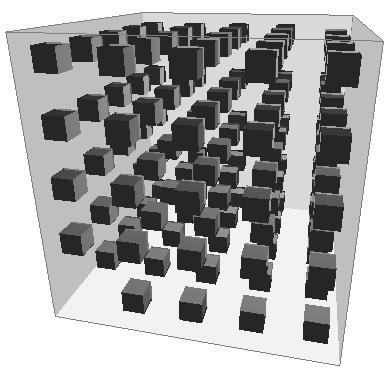} 
    \caption{Example of 3D clutter workspace}
    \label{fig:3DC} 
    \vspace{-4mm}
  \end{minipage}  
\end{figure}

To generate different workspace for training and testing, a number of quadrilateral blocks were placed in the operating region of $31 \times 31$ for 2D simple workspaces (2DS), $41 \times 41 \times 6$ for 3D office-like workspaces (3DO), and $11 \times 11 \times 11$ for 3D clutter workspaces (3DC). The placement of these blocks was randomly chosen in the operating region. Each random placement of the obstacle blocks leads to a different workspace. 
For each environment, the locations and numbers and types of obstacles are generated randomly. First, the number of obstacles is chosen randomly in a pre-defined range. Then for each obstacle in that environment, the type is chosen among the pre-defined types (small box, big box,..) and its location is also randomly determined.
Currently, we do not consider any rotation of the blocks, i.e., they are axis aligned.
Figure \ref{fig:2D}, \ref{fig:3DO}, and \ref{fig:3DC} show some examples of our workspaces.


We represent the workspaces as an occupancy grid with values of $-1$ and $1$, in which $1$ means there is an obstacle at that point, and $-1$ means the point is in free space. 
The input to the encoder is an occupancy grid of size $n \times m$ for 2D and $n \times m \times k$ for 3D, where $n$, $m$, and $k$ are the number of points along each dimension of the workspace.

We use multiple workspace representations to train and test our CAE. 
The training data set of the 2D workspaces comprised of 30 workspaces, and for testing, two types of test data sets were created to evaluate the proposed method. The first test data set comprised of the 30 workspaces used in training, and the second test data set comprised of 10 previously unseen workspaces, i.e., 10 workspaces not from the training set. There are around $3$ to $5$ obstacle blocks with different shapes and sizes in each workspace.

The training data set of the 3D office-like workspaces comprised of 100 workspaces, and for testing 100 known workspaces and 20 previously unseen workspaces with $25$ to $30$ obstacle blocks in each workspace were used.

The training data set of the 3D clutter workspaces comprised of 50 workspaces, and for testing 30 known workspaces and 10 previously unseen workspaces with $110$ to $125$ same shape and size obstacle blocks in each workspace were used.


To test the performance of our MLP, we generate multiple different workspaces with different starts and goals. For 2D simple and 3D office-like, we randomly generate 10 different workspaces with 10 different start and goal positions for each of them; and 100 3D clutter workspaces with 1 different start and goal positions for each of them to test the performance of our proposed framework in terms of time.

\subsubsection{Sample Data}

For the 2D simple workspace, we use a point-mass robot with 2 degrees of freedom (DoF) that represents its $\{x, y\}$ position. The point robot is (0.5x0.5) in a (31x31) environment.
For the 3D office-like, we use KuKa robot\cite{kuka-youbot}, which has 5 joints, with a total of 8 DoF. The 8 DoFs represent the robot's $\{x,y\}$ position, rotation, and each joint's rotation. The Kuka robot is the size of (1.5x1x1) in a (41x41x6) environment.
For the 3D clutter workspaces, we use box robots with 7 DoF and 9 DoF. The DoFs represent the robot's $\{x, y, z\}$ position, $\{\alpha, \beta, \gamma\}$ rotation, and each joint's rotation. 
The box robots consist of 2 types of links: a small link of (0.1x0.1x0.1) and a big link of (0.4x0.1x0.1) in a (11x11x11) environment. The 7D box robot has 1 joint for 2 links, and the 9D box robot has 3 joints for 2 small links and 2 big links.
We generate 100 samples for each 2D workspace and 200 samples for each 3D workspace for training and testing our MLP separately. This is evaluated as seen scenarios. Those samples are approximately distributed 50/50 between valid and invalid samples. 

Since our approach mainly focuses on verifying the sample's validity, we incorporate it into the standard PRM approaches using algorithm \ref{algo} to test its effectiveness. 
However, instead of having a large set of samples $S'$, we provide a small set of samples $S'$ and let the algorithm decide whether those supplied samples are enough or there is a need to generate more samples. We think that providing the algorithm with too many samples would negatively impact the performance of neighborhood finding routines in sampling-based planners. Thus, we generate 200 samples regardless of their validity for each 2D simple workspace, 2000 samples for each 3D office-like workspace, and 400 samples for each 3D clutter workspace. This is evaluated as unseen scenarios

\subsection{Model Architecture}
\subsubsection{Contractive AutoEncoder}

Since the structure of the decoders is the inverse of the encoder, we only describe the structure of the encoder. 

For the 2D workspaces, the encoding function $f(x, W^e)$ and decoding function $g(f(x,W^e), W^d)$ consist of five linear layers and one output layer, where each layer uses Parametric Rectified Linear Unit (PReLU)\cite{trottier2017parametric} as its activation function, for loss function, we use mean square error with weight-decay.
The layers 1 to 5 transforms the input vector $31 \times 31$ to 512, 256, 128, 64, and 32 hidden units, respectively. The output layer takes 32 units as input and outputs 12 units. Hence the workspace representation in the latent space is a vector of size 12. Then, those 12 units are used by the decoder to reconstruct the $31 \times 31$ space.

For the 3D office-like workspaces, the encoding function $f(x,W^e)$ and decoding function $g(f(x,W^e),W^d)$ are the same as in 2D simple workspace but consist of seven linear layers and one output layer. Layers one to seven transform the input vectors to 5043, 3125, 1600, 800, 400, 200, 100 hidden units, respectively. The output layer takes the 100 units from the seventh layer and transforms them into 50 units. 
For the 3D clutter workspaces, layers one to six transform the input vectors to 1000, 800, 600, 400, 200, 100 hidden units, the output layer transforms the input 100 units to output 50 units.

\subsubsection{Multi-layer Perceptrons}
The input is given by concatenating the encoded workspace’s representation $Z$, and the DoF for each robot from a given state. Each of the layers is a sandwich of a linear layer, a Parametric Rectified Linear Unit (PReLU), and Dropout ($p$).

For the 2D workspaces, MLP is a 4-layer neural network. Layers one and two map the input vectors to 6, 4 units, respectively. The output layer takes the 4 units and transforms them into the validity of each sample, which is either valid or invalid.
For the 3D workspaces, MLP is a 7-layer neural network.  Layers one to six map the input vectors to 50, 40, 30, 20, 10, 5 units, respectively. The output layer takes the 5 units from the sixth layer and transforms them into the validity of each sample.

\subsubsection{Parameters}
To train the neural models CAE and MLP for both 2D and 3D workspace, the Adagrad\cite{duchi2011adaptive} optimizer was used with the learning rate of $0.1$.
The Dropout probability $p$ and penalizing term $\lambda$ were set to $0.5$ and $0.001$, respectively.

\section{RESULTS}

This section presents the results of the framework for the motion planning of a point robot in the 2D workspaces and the KuKa robot and a box robot in 3D workspaces.

\begin{wraptable}{r}{1.55in}
\protect\scriptsize
\centering
\begin{tabular}{| c | c | c |}
 \hline
   & Seen  & Unseen \\ [0.5ex]
 \hline 
 2DS & 100\% & 100\% \\
 3DO & 96\% & 94\% \\
 3DC & 98\%  & 95\% \\
  \hline
 \end{tabular}
\caption{Performance of CAE}
\label{table:CAE}
\vspace{-4mm}
\end{wraptable}

Table \ref{table:CAE} shows the performance for the CAE. For 2DS, the average accuracy of our CAE is $100\%$ with the variance of $0.4\%$ for both already seen workspaces and unseen workspaces.
For 3DO, the average accuracy of our CAE is $96\%$ with the variance of $3\%$ for already seen workspaces and the average accuracy is $94\%$ with the variance of $5\%$ for unseen workspaces. The accuracy and variance are $98\%$ and $1\%$ and $95\%$ and $3\%$ for seen and unseen 3D clutter, respectively.
Overall, the accuracy of our CAE is excellent. Thus, we are able to learn the underlying structure of the workspaces quite well.

\begin{table}[h]
\centering
\begin{tabular}{|l|l|l|l|l|l|l|}
\hline
\multirow{2}{*}{} & \multicolumn{3}{c|}{Seen} & \multicolumn{3}{c|}{Unseen} \\ \cline{2-7} 
 & Acc & TPR & TNR & Acc & TPR & TNR \\ \hline
2DS & 92\% & 96\% & 88\% & 88\% & 90\% & 85\% \\ \hline
3DO & 76\% & 88\% & 64\% & 73\% & 85\% & 60\% \\ \hline
3DC-7 & 75\% & 87\% & 64\% & 71\% & 85\% & 61\% \\ \hline
3DC-9 & 73\% & 82\% & 66\% & 72\% & 82\% & 62\% \\ \hline
\end{tabular}

\caption{Accuracy, True Positive Rate, and True Negative Rate of MLP}
\label{table:MLP}
\vspace{-4mm}
\end{table}

Table \ref{table:MLP} shows the accuracy, true positive rate (TPR), and true negative rate (TNR) for the MLP, with false negative rate = 1–TPR and false positive rate = 1–TNR. For 2D workspaces, the average accuracy of our MLP is $92\%$ with the variance of $2\%$ for already seen workspaces and the average accuracy is $88\%$ with the variance of $4\%$ for unseen workspaces. For 3DO workspace, the average accuracy of our MLP is $76\%$ with the variance of $6\%$ for already seen workspaces, and the average accuracy is $73\%$ with the variance of $8\%$ for unseen workspaces. The average accuracy is $73\%$ with $5\%$ variance (seen), and $71\%$ with $9\%$ variance (unseen) for 3DC-7.  The average accuracy is $73\%$ with $6\%$ variance (seen), and $72\%$ with $10\%$ variance (unseen) for 3DC-9. Additionally, the true positive rates and true negative rates are quite good for our MLP, thus most of the valid samples and invalid samples are successfully validated by MLP. Overall, the accuracy of our MLP is acceptable. 
%

\begin{table}[h]
\begin{center}
\begin{tabular}{|l|l|l|l|l|l|l|l|l|}
\hline
\multirow{2}{*}{} & \multicolumn{4}{c|}{Sampling Time} & \multicolumn{4}{c|}{Total Time} \\ \cline{2-9} 
 & BS & OB & G & BT & BS & OB & G & BT \\ \hline
2DS & 53\% & 57\% & 63\% & 74\% & 17\%  & 18\%   & 11\% & 8\% \\ \hline
3DO & 45\% & 58\% & 41\% & 56\%  & 1\%  & 9\% & 10\% &  -8\%\\ \hline
3DC-7 & 49\% & 58\% & 51\% & 61\% & -2\% & 14\% & -9\% & 54\% \\ \hline
3DC-9 & 39\% & 59\% & 48\% & 62\% & 18\% & 20\% & 13\% &  37\%\\ \hline
\end{tabular}
\end{center}
\caption{Average time improvement ratio with and without applying our proposed framework for sampling time and total time.}
\label{table:tab3}
\vspace{-4mm}
\end{table}

\begin{figure}[!tbp]
  \centering
  \begin{minipage}[b]{0.5\textwidth}
    \centering
    \includegraphics[width=\textwidth, height = 90pt]{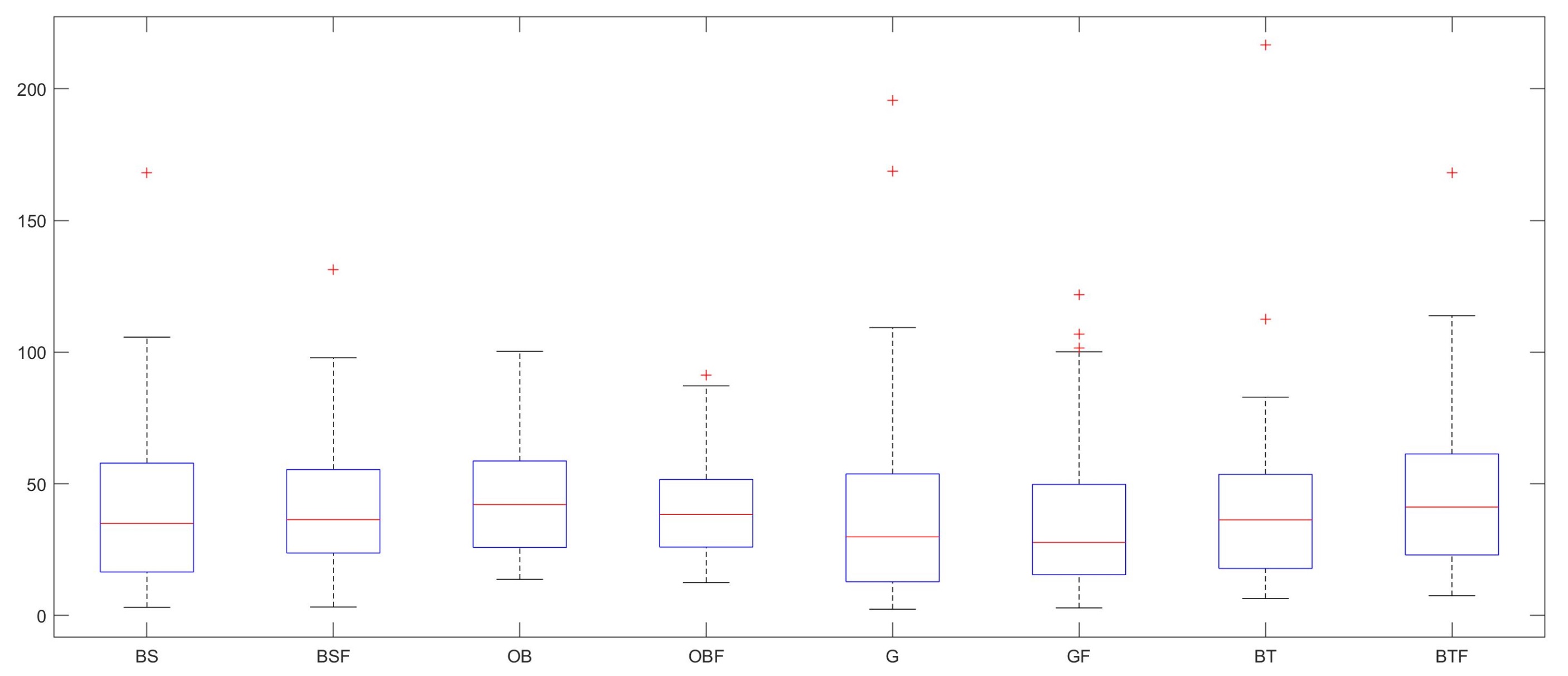} 
    \caption{Time performance with and without applying our method in 3DO workspaces}
    \label{fig:3DO-8} 
  \end{minipage}
  \hfill
  \begin{minipage}[b]{0.5\textwidth}
    \centering
    \includegraphics[width=\textwidth, height = 90pt]{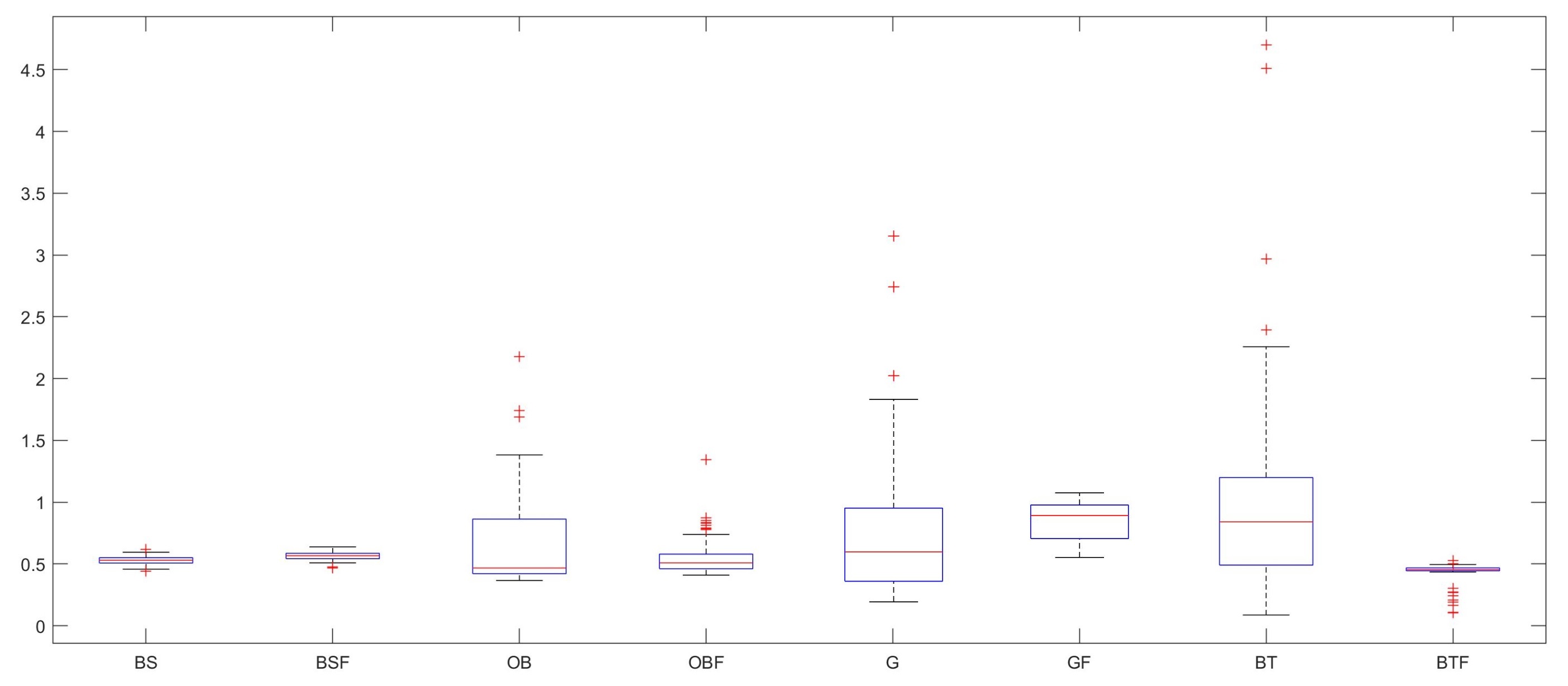} 
    \caption{Time performance with and without applying our method in 3DC-7 workspaces}
    \label{fig:3DC-7} 
  \end{minipage}
  \hfill
  \begin{minipage}[b]{0.5\textwidth}
    \centering
    \includegraphics[width=\textwidth, height = 90pt]{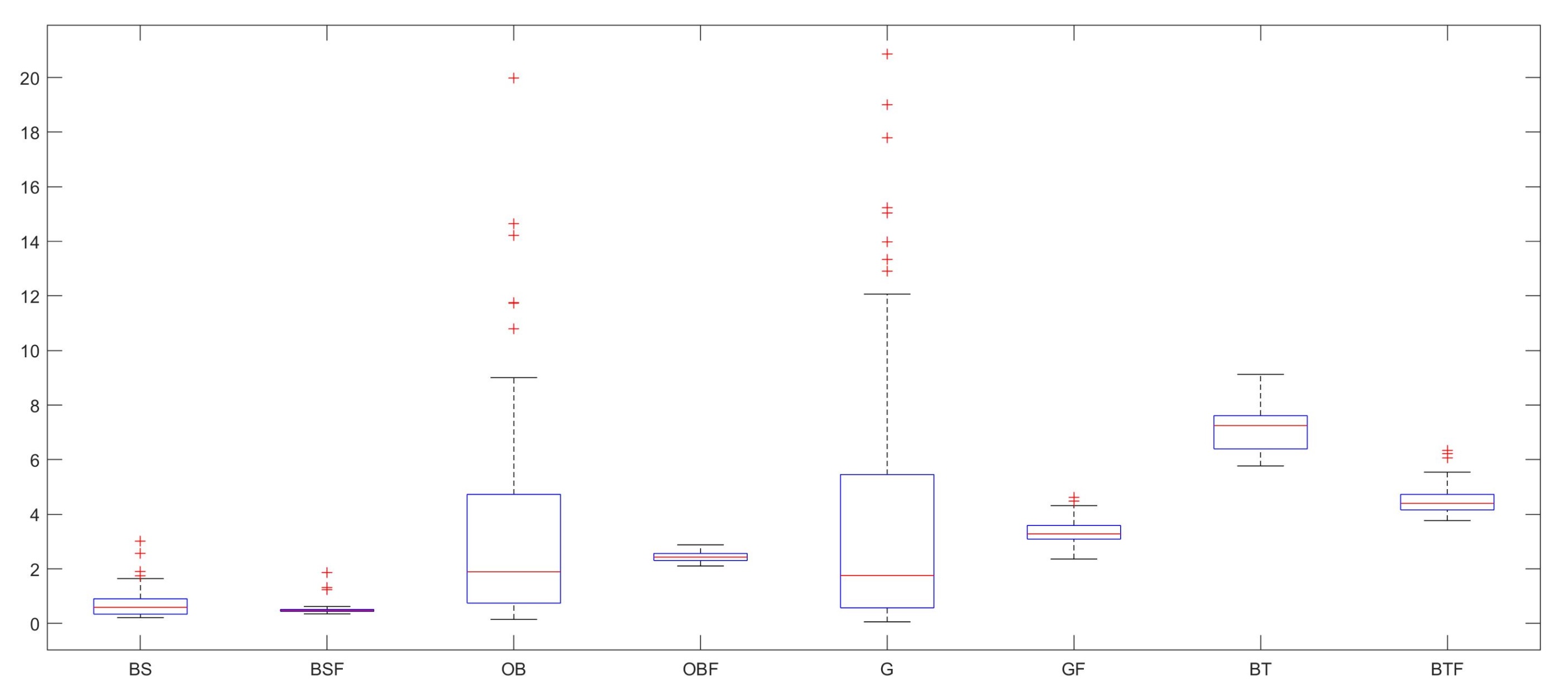} 
    \caption{Time performance with and without applying our method in 3DC-9 workspaces}
    \label{fig:3DC-9} 
    \vspace{-4mm}
  \end{minipage}  
\end{figure}






Table \ref{table:tab3} show the average time improvement ratio comparison for sampling time and total time with and without applying our method with Basic PRM (BS), Obstacle Based PRM (OB), Gaussian (G), and Bridge Test (BT) for solving motion planning problems.

In 2D workspaces, the average time to validate samples without and with using our method are: $0.0004 \pm  0.0002$ and  $0.0002 \pm 0.0002$ for BS, $0.05 \pm 0.01$ and $0.03 \pm 0.01$ for OB, $0.04 \pm 0.01$ and $0.02 \pm 0.01$ for G, $0.26 \pm 0.23$ and $0.11 \pm 0.03$ for BT.
There are significant time improvement when using our method for classify valid samples.
Moreover, the average total execution time for each query without and with using our method are: $0.12 \pm  0.01$ and  $0.10 \pm 0.01$ for BS, $0.38 \pm 0.05$ and $0.31 \pm 0.03$ for OB, $0.09 \pm 0.01$ and $0.09 \pm 0.01$ for G, $0.13 \pm 0.01$ and $0.13 \pm 0.01$ for BT.
Notable, our framework has good performance. 
Additionally, we notice there is a nearly $20\%$ improvement when applying our framework to OB.

In 3DO workspaces, the average time to validate samples without and with using our method are: $0.11 \pm  0.01$ and  $0.06 \pm 0.02$ for BS, $1.13 \pm 0.08$ and $0.56 \pm 0.25$ for OB, $0.21 \pm 0.04$ and $0.12 \pm 0.02$ for G, $0.76 \pm 0.13$ and $0.32 \pm 0.02$ for BT.
The time improvements when using our method for classifying valid samples are quite good.
Additionally, the average total execution time for each query without and with using our method are: $40.86 \pm  36.47$ and  $40.51 \pm 25.91$ for BS, $43.74 \pm 22.01$ and $39.88 \pm 18.72$ for OB, $42.32 \pm 41.99$ and $38.22 \pm 30.01$ for G, $40.40 \pm 33.92$ and $43.13 \pm 28.87$ for BT. There are improvements when using our framework with OB ($9\%$) and G ($10\%$), but there is a small improvement for BS ($1\%$) and a slow-down for BT ($8\%$). We suspect that it could be that the supplied samples are not good enough, thus there is the need for re-sampling. 

The average time to validate samples without and with using our method for 3DC-7 are: $0.007 \pm  0.003$ and  $0.003 \pm 0.001$ for BS, $0.20 \pm 0.03$ and $0.09 \pm 0.01$ for OB, $0.09 \pm 0.01$ and $0.04 \pm 0.02$ for G, $0.42 \pm 0.04$ and $0.17 \pm 0.02$ for BT; 
and for 3DC-9 are: $0.015 \pm  0.005$ and  $0.008 \pm 0.003$ for BS, $0.08 \pm 0.01$ and $0.04 \pm 0.01$ for OB, $0.04 \pm 0.01$ and $0.02 \pm 0.01$ for G, $0.10 \pm 0.02$ and $0.04 \pm 0.02$ for BT.
Overall, our method reduces the sampling classification time notably.
Moreover, the average total execution time for each query without and with using our method for 3DC-7  are: $0.53 \pm  0.03$ and  $0.54 \pm 0.03$ for BS, $0.65 \pm 0.35$ and $0.56 \pm 0.13$ for OB, $0.45 \pm 0.40$ and $0.49 \pm 0.12$ for G, $0.95 \pm 0.64$ and $0.44 \pm 0.05$ for BT; 
and for 3DC-9 are: $0.73 \pm 0.48$ and  $0.60 \pm 0.19$ for BS, $3.27 \pm 3.62$ and $2.60 \pm 0.19$ for OB, $3.90 \pm 4.73$ and $3.40 \pm 0.41$ for G, $7.10 \pm 0.77$ and $4.50 \pm 0.50$ for BT.
There are quite notable time improvements when using our framework to augment the motion planning algorithms for 3DC workspaces. In 3DC workspaces, the BT samplers perform the worst because of the open nature of our testing workspaces.




Figure \ref{fig:3DO-8}, \ref{fig:3DC-7}, \ref{fig:3DC-9} show the performance with (BSF, OBF, GF, BTF) and without (BS, OB, G, BT) applying our proposed framework in second for 3DO, 3DC-7, 3DC-9 workspace. Overall, there is a notable improvement in performance in most cases.
The standard deviations when using our framework decreased compared to not using it. Thus, our framework augments the algorithms perfectly and makes them more stable in solving queries reliably.
Although we only test our approach using fixed size workspaces, it is robust to varying sizes. 
By splitting the workspace into parts with a pre-defined size, the proposed approach would still be able to validate samples. 
Moreover, by adjusting the number of samples in the sampling validation phase and the number of samples generated by the auxiliary sampler, our method guarantees probabilistic completeness for any motion planning problem. Therefore, our approach is applicable to most of sampling based planners.

\section{Conclusion}

In this paper, we present a fast and efficient neural network framework for sampling-based motion planners. The framework consists of a Contractive AutoEncoder that encodes a point cloud representation of a robot's workspace into a latent feature space and a feed-forward neural network that takes that encoded workspace and robot configuration details to predict the validity of that configuration.
The main advantage of our framework is that we use an encoded workspace and neural network to reduce or even remove expensive collision detection operations. 
In this paper, we focus on the sampling phase of SBMP and how we can reduce number of collision checks made.
Our proposed method is quite robust such that it can be integrated into any sampling strategy. 

In the future, we plan to research models that increase the accuracy of our MLP by pre-processing a configuration's information before training.
Furthermore, since our method considers the environment’s and robot’s information when performing collision checks, we could train them on regions of the environments.
Thus, instead of using our method for the whole environment, an SBMP algorithm could use our method when it reaches certain potions of that environment. Additionally, we could apply this approach to other aspects of SBMP, e.g., edge prediction, and path validation.


\setcounter{page}{1}
\bibliographystyle{IEEEtran}
\bibliography{ref}

\end{document}